\documentclass[11pt]{article}
\usepackage[margin=1in]{geometry}
\usepackage{amsmath,amssymb,amsfonts}
\usepackage{booktabs}
\usepackage{graphicx}
\usepackage{hyperref}
\usepackage{natbib}
\usepackage{xcolor}
\usepackage{algorithm}
\usepackage{algpseudocode}
\usepackage{multirow}
\usepackage{cleveref}
\usepackage{enumitem}
\hypersetup{colorlinks=true,linkcolor=blue,citecolor=blue,urlcolor=blue}
\usepackage{titling}
\usepackage{authblk}
\usepackage[most]{tcolorbox}
\usepackage{xcolor}
\usepackage{tcolorbox}

\definecolor{examplebg}{RGB}{220,232,245}
\definecolor{exampleline}{RGB}{90,130,190}

\newtcolorbox{examplebox}[1]{
  enhanced,
  colback=examplebg,
  colframe=exampleline,
  boxrule=2pt,
  sharp corners,
  borderline north={1.2pt}{0pt}{exampleline},
  left=10pt,
  right=10pt,
  top=8pt,
  bottom=8pt,
  fonttitle=\large\bfseries,
  coltitle=white,
  title={#1}
}

\newcommand{\MR}{\textsc{MR}}
\newcommand{\Greedy}{\textsc{Greedy}}
\newcommand{\Random}{\textsc{Random}}
\newcommand{\Hardest}{\textsc{Hardest}}
\newcommand{\Easiest}{\textsc{Easiest}}
\newcommand{\Stratified}{\textsc{Stratified}}

\setcitestyle{round}

\pretitle{%
  \par
  \vspace*{0 em}
  \begingroup
  \centering
  \rule{\textwidth}{1.5pt}\par
  \vspace{0em}
  \LARGE\bfseries
}
\posttitle{%
  \par
  \vspace{0 em}
  \rule{\textwidth}{0.6pt}\par
  \endgroup
  \vspace{0 em}
}

\title{Efficient Benchmarking of AI Agents}

\author{Franck Ndzomga\footnote{\texttt{ndzomgafs@gmail.com}}}
\date{}

\begin{document}
\maketitle

\begin{abstract}
Evaluating AI agents on comprehensive benchmarks is expensive because each evaluation requires interactive rollouts with tool use and multi-step reasoning. We study whether small task subsets can preserve agent rankings at substantially lower cost. Unlike static language model benchmarks, agent evaluation is subject to scaffold-driven distribution shift, since performance depends on the framework wrapping the underlying model. Across eight benchmarks, 33 agent scaffolds, and 70+ model configurations, we find that absolute score prediction degrades under this shift, while rank-order prediction remains stable. Exploiting this asymmetry, we propose a simple optimization-free protocol: evaluate new agents only on tasks with intermediate historical pass rates (30--70\%). This mid-range difficulty filter, motivated by Item Response Theory, reduces the number of evaluation tasks by 44–70\% while maintaining high rank fidelity under scaffold and temporal shifts. It provides more reliable rankings than random sampling, which exhibits high variance across seeds, and outperforms greedy task selection under distribution shift. These results suggest that reliable leaderboard ranking does not require full-benchmark evaluation.
\end{abstract}

\noindent
\begin{figure}[H]
    \centering
    \includegraphics[width=0.81\textwidth]{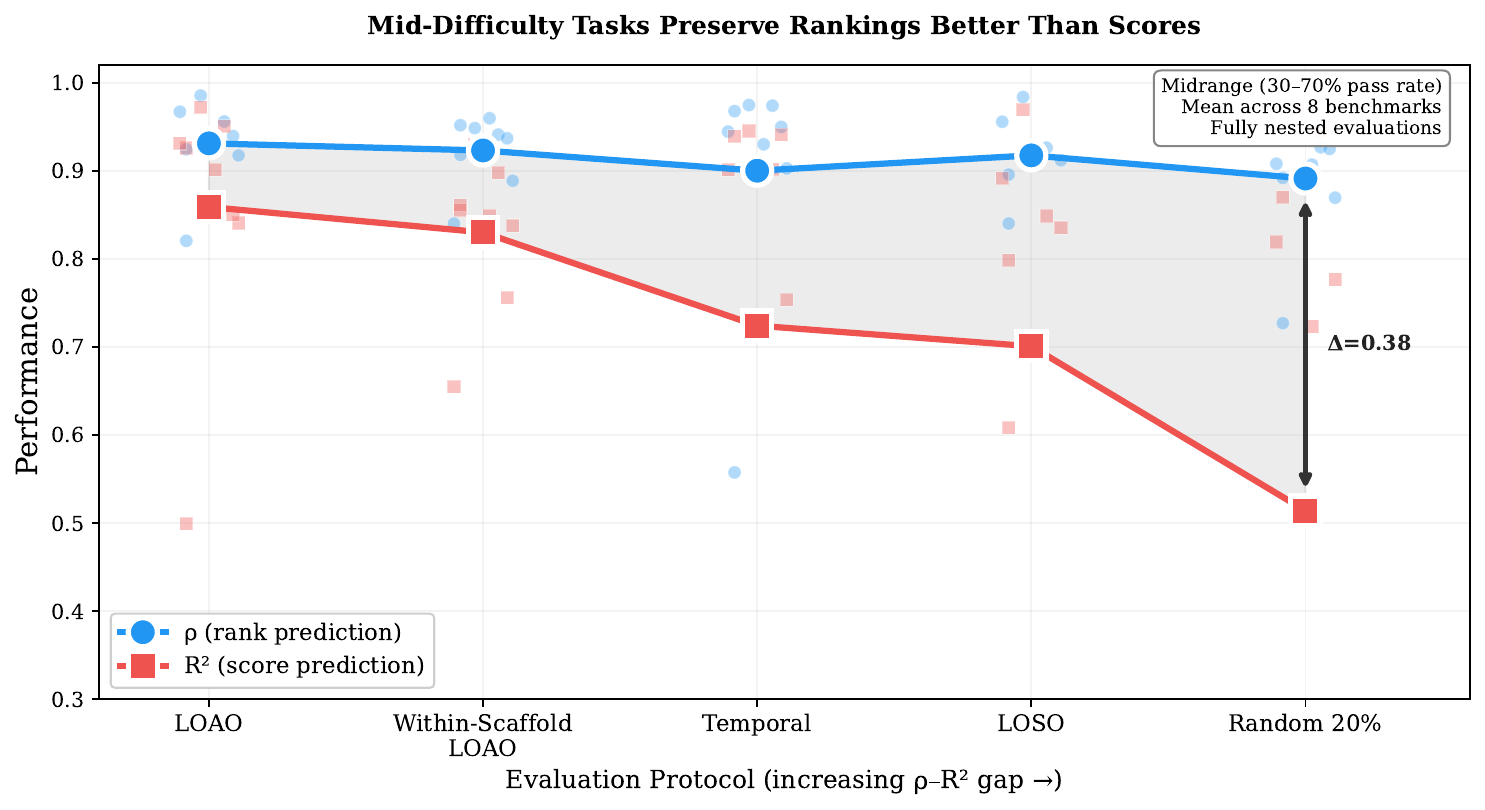}
    \caption{\textbf{The Robustness Gap.} Spearman $\rho$ (ranking) and $R^2$ (score prediction) across evaluation regimes. Ranking fidelity remains robust even as absolute score prediction rapidly degrades under temporal, scaffold and random shift.}
    \label{fig:robustness_gap}
\end{figure}

\section{Introduction}

Agent benchmarks are increasingly used to evaluate AI systems on complex, multi-step tasks requiring reasoning, tool use, and interaction with external environments. Running these benchmarks is expensive and creates a situation reminiscent of the compute divide in Machine Learning research \citep{besiroglu2024computedividemachinelearning}.
The Holistic Agent Leaderboard \citep{hal2026}---one of the most comprehensive standardized evaluation efforts for AI agents---required roughly \$40{,}000 to evaluate agents on nine benchmarks, despite considering at most two scaffolds per benchmark and only one run per scaffold–model configuration. Such costs create a barrier for independent researchers and small labs, and make statistically robust evaluation difficult in practice
\citep{gonzalez2025repetitionsmatterstrengtheningreliability}.

Prior work on benchmark reduction has shown that NLP and language-model evaluations can often be compressed to much smaller task subsets while preserving the ability to predict model scores
\citep{vivek2024anchorpointsbenchmarkingmodels,tinybench2024,
perlitz2024efficientbenchmarkinglanguagemodels,
subramani2025simbasimplifyingbenchmarkanalysis}.
Agent benchmarks, however, introduce a source of shift that static evaluations lack: performance depends not only on the underlying model but also on the \emph{scaffold}, the harness governing tool use, memory, retry logic, and execution flow. A reduced evaluation suite for agents must therefore generalize across scaffolds and over time, not only across models. The reduction problem is also more stringent: agent benchmarks typically contain dozens to hundreds of tasks rather than thousands, and each task
requires a full agent loop rather than a single prompt--response pass.

This paper studies whether the number of benchmark tasks can be significantly reduced while preserving the signal that agent leaderboards actually consume: \emph{rankings}.
We make three contributions:

\begin{enumerate}[leftmargin=*,itemsep=3pt]
\item We identify a robust \textbf{empirical asymmetry between ranking and score prediction under scaffold and temporal shift}, which makes benchmark reduction feasible in the agent setting.

\item We propose the \textbf{Mid-Range Difficulty Filter (\MR{})}, a deterministic, optimization-free task selection rule that retains tasks with pass rates between 30--70\%. The rule is motivated by Item Response Theory: tasks near 50\% pass rate carry the strongest discriminative information about latent agent ability.

\item We evaluate \MR{} against greedy, random, stratified, and extreme-difficulty baselines under five protocols of increasing distributional shift, using proper nested cross-validation throughout. \textbf{\MR{} achieves the best ranking fidelity across the eight benchmarks we study} while remaining stable across evaluation regimes, in contrast to the high-variance behavior of random and greedy selection.
\end{enumerate}

Together, these results support a practical conclusion: routine
leaderboard evaluation can default to reduced task suites, with full-benchmark runs reserved for initialization, drift monitoring, and major capability transitions.

\section{Background and Related Work}

A growing literature studies how to reduce the cost of benchmark evaluation without substantially changing benchmark scores or rankings. Most prior work focuses on NLP benchmarks or language model (LM) evaluations rather than agent benchmarks. For example, \cite{vivek2024anchorpointsbenchmarkingmodels} propose Anchor Point Selection, a method for identifying small representative subsets of benchmark examples that can reliably rank models and estimate instance-level behavior on the full dataset. Similarly, \cite{tinybench2024} shows that the performance of a LM in suites such as MMLU \citep{hendrycks2021measuring} and HELM \citep{liang2023holisticevaluationlanguagemodels} can be estimated from a much smaller set of carefully selected examples, which produces large reductions in the evaluation cost. \cite{perlitz2024efficientbenchmarkinglanguagemodels} studies the computation--reliability trade-off in HELM and proposes the Decision Impact on Reliability (DIoR) measure together with an evaluation algorithm that can substantially reduce the evaluation cost while preserving the reliability of the ranking. More recently, \cite{subramani2025simbasimplifyingbenchmarkanalysis} studied benchmark reduction from performance matrices alone, showing that small representative subsets of datasets can preserve model rankings and predict held-out performance on benchmarks such as HELM, MMLU and BigBenchLite \citep{srivastava2023beyond}.

A related line of work uses psychometric ideas, especially Item Response Theory (IRT) \citep{baker2001_irt}, to determine which benchmark items are the most informative \citep{GIGNAC2025101922}. \cite{rodriguez-etal-2021-evaluation} propose DAD (Difficulty and Ability Discriminating) leaderboards, which model latent item difficulty and model ability to analyze leaderboard reliability and identify informative or erroneous evaluation items. Applying the DAD framework to the SQuAD 2.0 leaderboard shows that IRT-based rankings are more stable than rankings based solely on average accuracy and can improve the reliability of evaluation when only a small subset of items is available.These approaches emphasize that benchmark items differ in difficulty and discrimination, and that benchmark rankings can change depending on which items are selected. Recent work extends this perspective to LM evaluations, for example by using richer IRT models to diagnose benchmark quality and construct smaller benchmark variants that better align with human preferences \citep{zhou2026lostbenchmarksrethinkinglarge}.

Our work is close in spirit to these papers, but differs in setting and objective. Prior benchmark reduction work primarily studies model evaluation on NLP or LM benchmarks. Agent benchmarks introduce an additional and central source of shift: performance depends not only on the underlying model, but also on the scaffold that governs tool use, memory, retry logic, and execution flow. As a result, reduced evaluations for agents must generalize not only across models, but also across scaffolds and over time. The reduction problem is also more stringent in the agent setting. In LM evaluations, reducing a benchmark often means moving from thousands of examples to a few hundred. Agent benchmarks, however, often contain only dozens or hundreds of tasks to begin with, and each task is much more expensive because it requires a full agent loop rather than a single prompt--response pass. To be genuinely cost-efficient, a reduced agent benchmark must therefore extract a strong signal from both fewer tasks and relatively few fully evaluated agents.

\section{Methods}

\subsection{Problem Setup}

Let us suppose that for every benchmark at our disposal, we have access to a per task performance matrix $\mathbf{X} \in [0,1]^{n \times m}$, where $n$ is the number of agents and $m$ is the number of tasks. Each entry $X_{ij}$ indicates whether agent~$i$ solved task~$j$. When each agent is evaluated once per task, entries are binary ($X_{ij} \in \{0,1\}$); when multiple trials are run, $X_{ij}$ is the fraction of trials in which agent~$i$ solved task~$j$. The full benchmark score for agent~$i$ is the mean across all tasks:

\begin{equation}
  y_i = \frac{1}{m} \sum_{j=1}^{m} X_{ij}
\end{equation}

Collecting these scores over all agents gives the benchmark score vector
\begin{equation}
    \mathbf{y} = (y_1,\dots,y_n)^\top.
\end{equation}

The goal of benchmark reduction is to identify a subset $S \subseteq \{1, \ldots, m\}$ with $|S| = k \ll m$ such that performance on $S$ alone predicts $\mathbf{y}$. Let $\mathbf{X}_S \in [0,1]^{n \times k}$ denote the submatrix of $\mathbf{X}$ restricted to the columns in $S$. We fit a Ridge regression on the selected columns, $\hat{\mathbf{y}} = \mathbf{X}_S \boldsymbol{\beta}$, where $\boldsymbol{\beta} \in
\mathbb{R}^k$ is the Ridge coefficient vector with regularization parameter $\alpha = 1.0$ throughout for score calibration. For rank prediction alone, an unweighted mean over selected tasks suffices.

A critical distinction in this work is between two prediction targets that impose different requirements on the task subset.

\paragraph{Score prediction.} We ask how accurately the subset predicts each agent's full benchmark score, measured by the coefficient of determination:

\begin{equation}
  R^2 = 1 - \frac{\sum_i (y_i - \hat{y}_i)^2}{\sum_i (y_i - \bar{y})^2}
\end{equation}

where $\hat{y}_i$ is the Ridge-predicted score for agent~$i$ and
$\bar{y} = \frac{1}{n}\sum_i y_i$ is the mean benchmark score across agents. High $R^2$ requires the subset to capture the absolute difficulty structure of the benchmark---not just which agents are better, but by how much (calibration). This is sensitive to any shift in the relationship between subset performance and full-benchmark performance.

\paragraph{Rank prediction.} We ask whether the subset preserves the ordering of agents, measured by two complementary rank correlations. Spearman's $\rho$ is the Pearson correlation computed on rank vectors:

\begin{equation}
  \rho = \frac{\sum_i (r_i - \bar{r})(s_i - \bar{s})}
              {\sqrt{\sum_i (r_i - \bar{r})^2}\;\sqrt{\sum_i (s_i - \bar{s})^2}}
\end{equation}

where $r_i$ and $s_i$ are the ranks of agent~$i$ in the predicted and actual orderings respectively (with ties assigned midranks), and $\bar{r}$, $\bar{s}$ are their means. It captures overall monotonic association and is sensitive to large rank displacements.

Kendall's $\tau$ counts concordant and discordant pairs, adjusted for ties (we use the $\tau_b$ variant throughout and write $\tau$ for brevity):

\begin{equation}
  \tau = \frac{C - D}{\sqrt{(n_0 - n_1)(n_0 - n_2)}}
\end{equation}

where a pair $(i,j)$ is \emph{concordant} if $\hat{y}_i > \hat{y}_j$ and $y_i > y_j$ (or both inequalities reversed) and \emph{discordant} if the inequalities disagree; $C$ and $D$ are the counts of concordant and discordant pairs, $n_0 = \binom{n}{2}$ is the total number of pairs, and $n_1$, $n_2$ are the number of pairs tied in the predicted and actual rankings respectively. Kendall's $\tau$ has a direct probabilistic interpretation: $(\tau + 1)/2$ is the probability that a randomly chosen pair of agents is ranked correctly. For instance, $\tau = 0.80$ means that 90\% of all pairwise comparisons agree with the full
benchmark. Because $\tau$ counts pairwise errors rather than squared rank displacements, it provides a stricter test of ranking fidelity than $\rho$, which tends to be numerically higher for the same underlying agreement. We report both throughout: $\rho$ for comparability with prior work, $\tau$ for interpretability.

These two targets can diverge substantially. A task subset can produce poorly calibrated score predictions ($R^2 \ll 1$) while still correctly identifying which agents outperform which ($\rho \approx 1$, $\tau \approx 1$). As we show in Section~\ref{sec:results}, this
divergence is not merely theoretical: under scaffold-induced distribution shift, $R^2$ collapses while both rank metrics remain high. This asymmetry is the key to cost-efficient agent evaluation, because most practical use cases---model selection, scaffold selection, leaderboard ranking---require only rank prediction.

To achieve benchmark reduction, one needs a \emph{task selection method}: a function $\sigma$ that maps the entire task set $\{1,\ldots,m\}$ (and, optionally, historical performance data) to a subset $S = \sigma(\{1,\ldots,m\}) \subseteq \{1,\ldots,m\}$ with
$|S| = k \ll m$.

\subsection{Data}
\label{sec:data}

We study eight agent benchmarks: Terminal-Bench~2.0 \citep{terminalbench2026}, a benchmark of AI agents in terminal environments with rich temporal structure, and seven benchmarks from the Holistic Agent Leaderboard~(HAL), spanning diverse task domains but with smaller agent populations.

\paragraph{Terminal-Bench 2.0.} Terminal-Bench evaluates AI agents on 89~tasks, each attempted 5~times per agent, yielding a fractional success rate per cell. At the time we scraped it, the leaderboard contained 101~agents drawn from 23~distinct agent scaffolds (OpenHands \citep{wang2025openhandsopenplatformai}, Codex~CLI \citep{openai_codex_cli_2025}, Aider \citep{aider2024}, and various custom implementations). Terminal-Bench has a natural temporal structure: 50~agents were present at launch in October~2025, and 51 were added over the following 3.5~months across 17~scaffold types not represented in the original batch. This makes Terminal-Bench uniquely suited for evaluating whether the task selection part of the benchmark reduction process generalizes to novel agent architectures.

\paragraph{Holistic Agent Leaderboard.} HAL \citep{hal2026} contains results from nine benchmarks spanning coding (SWE-bench Verified \citep{jimenez2024swebenchlanguagemodelsresolve}, CoreBench Hard \citep{siegel2024corebenchfosteringcredibilitypublished}, USACO \citep{shi2024languagemodelssolveolympiad}, SciCode \citep{tian2024scicoderesearchcodingbenchmark}, ScienceAgentBench \citep{chen2025scienceagentbenchrigorousassessmentlanguage}), web navigation (Online Mind2Web \citep{xue2025illusionprogressassessingcurrent}), general assistance (GAIA \citep{gaia2024}, AssistantBench \citep{yoran2024assistantbenchwebagentssolve}), and customer service (TAU-bench Airline \citep{yao2025taubench}). ScienceAgentBench and AssistantBench are excluded from our analysis because they do not provide the clean binary per-task success signals necessary for task selection and ranking prediction. TAU-bench uses reward-thresholding rather than a native binary success report per task: a task is considered successful if the reward is strictly positive, and overall accuracy corresponds to the average reward across tasks. We found that reward-thresholding still provides an exploitable per-task signal, so we retained TAU-bench in our analysis. Table~\ref{tab:benchmarks} summarizes the key statistics of each benchmark at the time of data acquisition.

\begin{table}[H]
\centering
\small
\begin{tabular}{lrrrr}
\toprule
Benchmark & Agents & Tasks & Scaffolds & Avg task $\rho$ \\
\midrule
CoreBench Hard & 38 & 45 & 2 & 0.10 \\
GAIA & 32 & 165 & 2 & 0.15 \\
Mind2Web & 22 & 300 & 2 & 0.03 \\
SciCode & 24 & 65 & 2 & 0.12 \\
SWE-bench Verified & 28 & 50 & 2 & 0.29 \\
TAU-bench Airline & 26 & 50 & 2 & 0.08 \\
USACO & 13 & 307 & 2 & 0.17 \\
\midrule
Terminal-Bench 2.0 & 101 & 89 & 23 & 0.30 \\
\bottomrule
\end{tabular}
\caption{Benchmark statistics. \textbf{Avg task $\rho$}: mean pairwise Spearman correlation between task outcome vectors across agents; higher values indicate a more homogeneous benchmark.}
\label{tab:benchmarks}
\end{table}

Benchmarks from HAL differ from Terminal-Bench in three important respects. First, in most cases, they record a single trial per task, producing binary outcomes with no within-cell variance estimation; this makes the per task signal noisier. Second, each HAL benchmark includes only two scaffolds, limiting the diversity of the agent population. Third, agent sample sizes are modest ($n = 13$--$38$), constraining the reliability of any data-driven selection procedure.

The final column in Table~\ref{tab:benchmarks}, average task~$\rho$, reports the mean pairwise Spearman correlation between task outcome vectors. Higher values indicate that agent performance is consistent across tasks; lower values indicate more variable performance patterns.

\paragraph{Cost structure.} Table~\ref{tab:costs} reports per-run evaluation costs from HAL. A single SWE-bench Verified run costs a median of \$163, with per-task costs ranging from \$0.08 (DeepSeek~R1) to \$32.00 (Claude Opus 4.1 High) -- a spread of $400\times$ driven by model pricing and scaffold design. Evaluating one agent on all eight selected HAL benchmarks costs a median of roughly \$800 in API fees. The total cost across all 242 agent runs is approximately \$46{,}000, consistent with HAL's reported estimate of $\sim$\$40{,}000~\citep{hal2026} since additional runs were added between the publication of the paper and our data acquisition efforts. The total cost of one agent run depends on the benchmark, the scaffold, and the model used. State of the art models can push costs as high as \$2{,}829 on GAIA before accounting for caching benefits.

\begin{table}[H]
\centering
\small
\begin{tabular}{lrrrrrr}
\toprule
Benchmark & Tasks & Agents & Med.\ \$/run & Max \$/run & Med.\ \$/task & Range \$/task \\
\midrule
SWE-bench Verified & 50 & 33 & \$163 & \$1600 & \$3.26 & \$0.08--32.00 \\
CoreBench Hard & 45 & 45 & \$66 & \$510 & \$1.47 & \$0.05--11.33 \\
GAIA & 165 & 32 & \$140 & \$2829 & \$0.85 & \$0.05--17.14 \\
Mind2Web & 300 & 22 & \$276 & \$1610 & \$0.92 & \$0.02--5.37 \\
SciCode & 65 & 33 & \$67 & \$625 & \$1.03 & \$0.00--9.62 \\
TAU-bench Airline & 50 & 26 & \$22 & \$180 & \$0.44 & \$0.01--3.61 \\
USACO & 307 & 13 & \$56 & \$276 & \$0.18 & \$0.00--0.90 \\
\bottomrule
\end{tabular}
\caption{Per-run evaluation costs on HAL benchmarks. Costs reflect API fees and vary with model and scaffold.}
\label{tab:costs}
\end{table}

The Terminal-Bench 2.0 leaderboard does not offer the level of fine-grained details HAL offers to track the cost of each agent run.

\subsection{Task Selection Strategies}
\label{sec:strategies}

The challenge of agent benchmark reduction consists of finding a task selection strategy that allows us to run fewer tasks while still recovering agent rankings and benchmark scores. We evaluate three main task selection strategies: Mid-Range, Greedy, and Random selection.

\paragraph{Mid-Range Difficulty Filter (\MR{}).} We select all tasks whose pass rate falls in the interval $[0.30, 0.70]$. The choice of band is motivated by Item Response Theory \citep{baker2001_irt}. Let $p$ denote the probability that an agent solves a task. Under a Bernoulli response model with logistic link, the Fisher information about the latent ability parameter $\theta$ is $I(\theta) = p(1-p)$. This quantity is maximized at $p = 0.5$ and falls to half its maximum at $p \approx 0.146$ and $p \approx 0.854$.

We target a benchmark reduction of roughly $50\%$ as a practical efficiency requirement. The $30$--$70\%$ band provides a simple fixed rule that is substantially narrower than the full high-information region ($15$--$85\%$), while still retaining moderately difficult tasks.

We apply the mid-range filtering protocol only to benchmarks with a sufficiently populated intermediate-difficulty band. In our data, SciCode contains only four tasks in the 30--70\% pass-rate interval, making the protocol unstable. We therefore exclude it from the main \MR{} evaluation results, and retain it only in the per-benchmark breakdown to illustrate a case where the mid-range protocol is not appropriate.

\paragraph{Greedy Forward Selection (\Greedy{}-$k$).} Starting from an empty set, we iteratively add the task that maximizes cross-validated leave-one-agent-out $R^2$ under Ridge regression \citep{Hoerl01021970}, using the hat-matrix shortcut for efficient computation. For fair comparison with MR, we use $k$ tasks, where $k$ equals the number of mid-range tasks for that evaluation fold. Specifically, in each evaluation fold, we first compute the MR band on the training agents to obtain a fold-specific budget $k_i$, then evaluate Greedy (and all other baselines) at exactly that budget. This fully nested, matched-budget design ensures no information leakage and enables direct comparison across methods.

\paragraph{Random Sampling (\Random{}-$k$).} We sample $k$ tasks uniformly at random, where $k$ is matched to the MR pool size in each evaluation fold for fair comparison. We use 100 random seeds; a meta-bootstrap analysis confirms this yields stable estimates across all benchmarks (variance of the estimated mean and standard deviation below $2 \times 10^{-5}$). We report mean $\pm$ standard deviation across seeds. This baseline tests whether the identity of selected tasks matters, or whether any sufficiently large random subset suffices.

\paragraph{Other baselines.} We also evaluate \Hardest{}-$k$ (the $k$ tasks with lowest pass rate), \Easiest{}-$k$ (the $k$ tasks with highest pass rate), and \Stratified{}-$k$ (uniform sampling across difficulty deciles). All baselines use the same fold-specific budget $k$ as MR.

\subsection{Evaluation Protocols}
\label{sec:protocols}

All evaluations use proper nested cross-validation (CV): task selection occurs inside the CV loop, so the test agent's data is never used during task selection. This prevents the optimistic bias that arises when tasks are selected on the full dataset before splitting.

We evaluate under five protocols, ordered by increasing distributional shift:

\begin{enumerate}[leftmargin=*,itemsep=3pt]
\item \textbf{Within-Scaffold LOAO.} Leave-one-agent-out, restricted to agents sharing the same scaffold. On Terminal-Bench, we evaluate on the three scaffolds with $\geq$10 agent runs (Terminus~2: 30, Mini-SWE-Agent: 13, OpenHands: 12). On HAL, each scaffold typically contains 10--20 agent runs, and we discard the ones that do not.

\item \textbf{LOAO.} Leave-one-agent-out across all agents regardless of scaffold. Tasks are reselected for each fold.

\item \textbf{Random 80/20 splits.} Random partition of agents into 80\% train, 20\% test ($\times$100 seeds).

\item \textbf{LOSO.} Leave-one-scaffold-out: train on all agents from other scaffolds, test on the held-out scaffold. Tasks selected on training scaffolds only.

\item \textbf{Temporal expanding window.} Train on all agents submitted before time $t$, predict the agent submitted at time $t$. Tasks selected on the training set at each step. This mirrors real leaderboard operation.
\end{enumerate}

We report Spearman $\rho$, Kendall $\tau$, and $R^2$.

\section{Results}
\label{sec:results}

\subsection{The $\rho$--$R^2$ Divergence: Rank Prediction is Preserved Even When Score Prediction Collapses}

By running 5 evaluation protocols across 8 agent benchmarks and 6 task selection strategies, we show empirically that leaderboard rankings are preserved better than absolute scores.

Rank prediction is stable across protocols ($\rho = 0.90$--$0.96$), while score prediction drops from $R^2 = 0.90$ to $R^2 = 0.54$ under random splits and $R^2 = 0.65$ under scaffold shift. On individual benchmarks, the divergence is more extreme: $R^2$ goes negative on Online Mind2Web under LOSO while $\rho$ remains above 0.90. The temporal expansion window protocol shows high $\rho$ (0.90) because the growing training set offsets the distributional shift. It is the protocol that mirrors the most real leaderboard operation.

Table~\ref{tab:divergence} reports the central finding: under proper nested cross-validation, Spearman~$\rho$ remains stable across all evaluation protocols while $R^2$ degrades substantially. This empirical reality can be exploited to reduce the cost of AI agent evaluation. It means that we need not run full benchmark suites to be able to compare agents against each other and rank them. The worst Kendall's $\tau$ among optimal task selection strategies is $0.80$ (LOSO), meaning that $90\%$ of agent pairs are ranked correctly: for every $100$ pairwise comparisons, only $10$ are inverted.

\begin{table}[H]
\centering
\begin{tabular}{l cc cc cc c}
\toprule
& \multicolumn{2}{c}{$\rho$} & \multicolumn{2}{c}{$\tau$} & \multicolumn{2}{c}{$R^2$} & \\
\cmidrule(lr){2-3} \cmidrule(lr){4-5} \cmidrule(lr){6-7}
Protocol & Best & Avg & Best & Avg & Best & Avg & Benchmarks \\
\midrule
LOAO & 0.95 & 0.91 & 0.85 & 0.79 & 0.90 & 0.74 & 8 \\
Within-Scaffold LOAO & 0.96 & 0.86 & 0.87 & 0.74 & 0.89 & 0.65 & 8 \\
Temporal & 0.90 & 0.85 & 0.81 & 0.73 & 0.71 & 0.52 & 8 \\
LOSO & 0.92 & 0.87 & 0.80 & 0.73 & 0.65 & 0.50 & 7 \\
Random 20\% & 0.90 & 0.86 & 0.84 & 0.79 & 0.54 & 0.46 & 8 \\
\bottomrule
\end{tabular}
\caption{Rank prediction ($\rho$, $\tau$) vs.\ score prediction ($R^2$) across evaluation protocols. ''Best'' selects the top-performing task selection method per benchmark; ''Avg'' averages across methods.}
\label{tab:divergence}
\end{table}

Agent benchmarks are often used to estimate absolute capability. In practice, even a full run of a benchmark suite merely provides a biased estimate of the capability we are trying to measure. The estimate is biased by distributional limitations (benchmark tasks are a convenience sample, not a representative draw from the space of all relevant challenges), construct validity failures \citep{zhu2025establishingbestpracticesbuilding} (seven of ten widely used agent benchmarks exhibit either task validity failures, where a trivial agent can pass without possessing the target capability, or outcome validity failures, where the grader awards credit for incorrect completions), and scaffold confounding (performance depends jointly on the model and the scaffold wrapping it, so absolute scores conflate capability with confounding engineering choices). These biases are structural: they affect the expectation of the score, not just its variance, and cannot be eliminated by running more tasks or more agents. Focusing on rank prediction rather than score reconstruction is therefore defensible on epistemic grounds, not merely practical ones: absolute scores are already systematically misleading before any reduction is applied, and rankings inherit less of this structural bias because affine distortions to the score scale leave orderings intact. In other words, ranking agents on a benchmark is more defensible than estimating their absolute capabilities, since it requires no assumption that the benchmark reliably measures the underlying capability it is intended to assess.

\subsection{Mid-Range Task Selection Is the Best Strategy Across Benchmarks and Evaluation Protocols}
The Mid-Range selection strategy preserves
ranking better than all the other strategies evaluated. Although most strategies can achieve high peak performance ($\rho \approx 0.99$) under optimal conditions, their lower bounds reveal severe vulnerabilities to distribution changes. Random and Greedy can have a Spearman~$\rho$ as low as 0.54 and 0.56 respectively.

\begin{figure}[htb]
    \centering
    \includegraphics[width=0.75\textwidth]{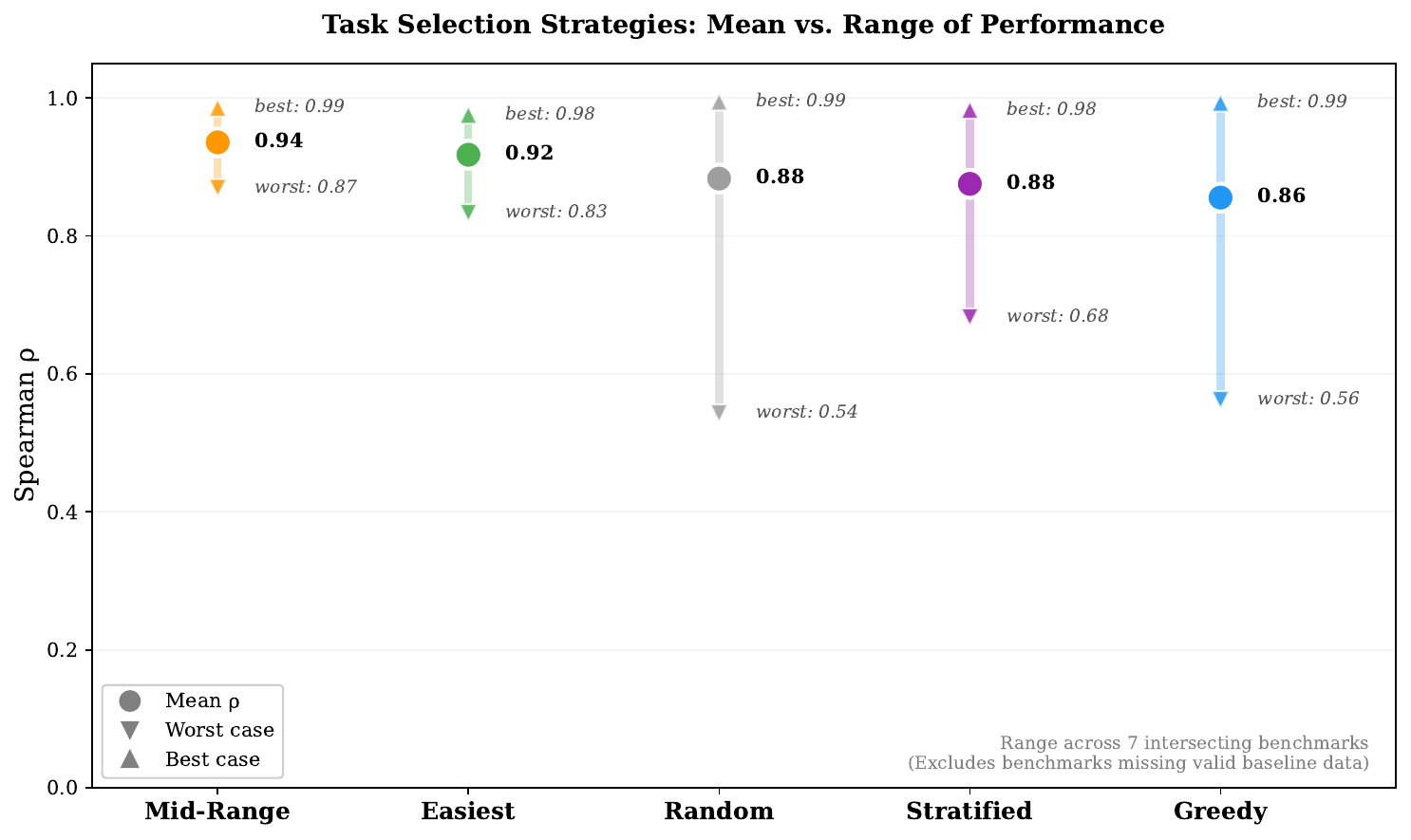}
    \caption{\textbf{Range of Performance by Selection Strategy.} Mean, best-case, and worst-case Spearman $\rho$ across overlapping benchmarks and evaluation protocols. Mid-Range task selection offers the overall best ranking preservation across benchmarks.}
    \label{fig:selection_robustness}
\end{figure}

Mid-Range task selection (\MR{}) maintains the highest mean ($\rho=0.94$) while providing a robust safety net (worst-case $\rho=0.87$), avoiding the catastrophic degradation seen in Greedy and Random selection, as presented in Figure~\ref{fig:selection_robustness}.

\begin{figure}[htb]
    \centering
    \includegraphics[width=0.8\textwidth]{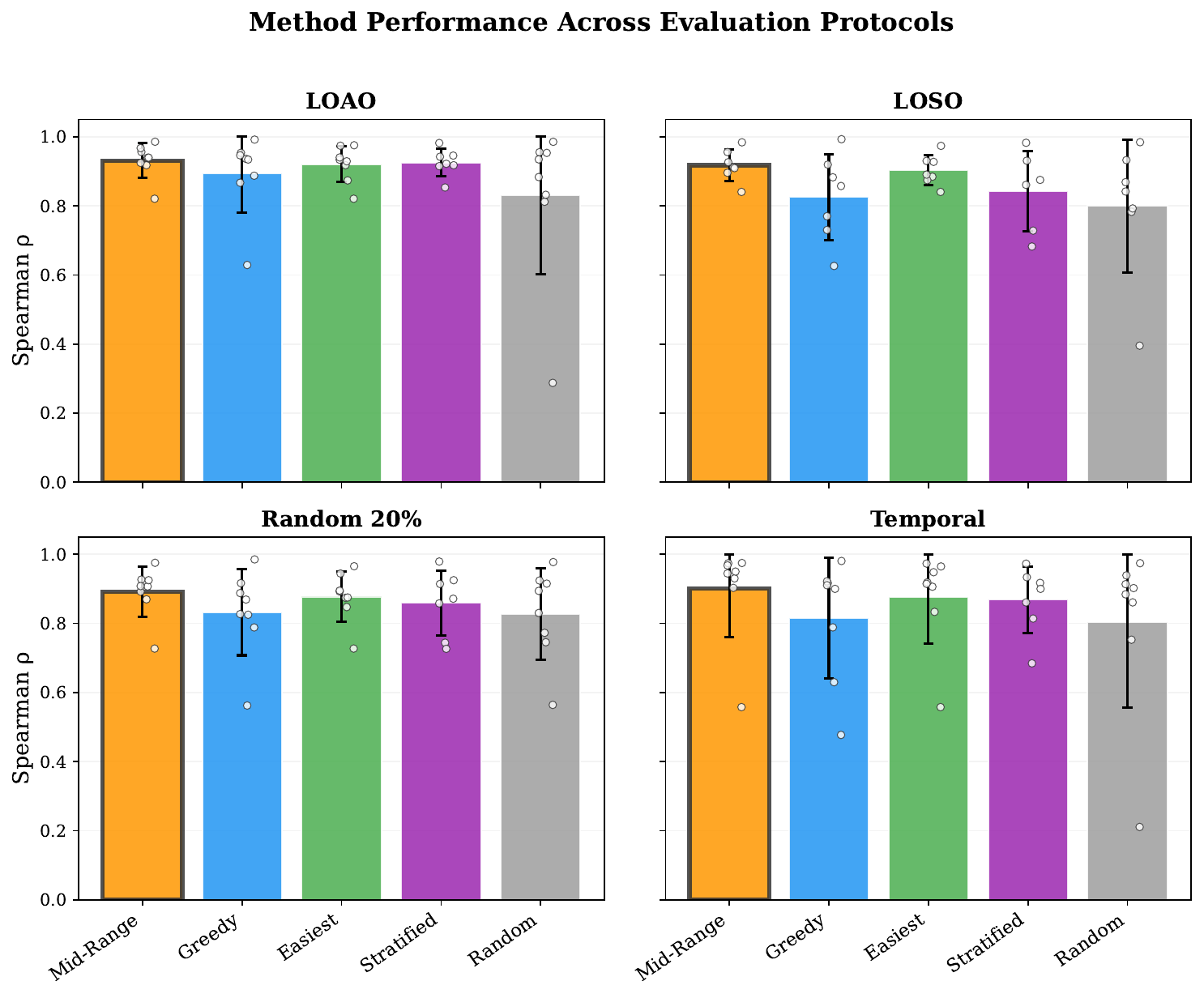}
    \caption{\textbf{Performance Stability Across Distribution Shifts.} Average Spearman $\rho$ by task selection strategy, disaggregated across four evaluation protocols (LOAO, LOSO, Random 20\%, and Temporal shift). Mid-Range selection maintains reliable rankings ($>0.85$ mean $\rho$) regardless of the evaluation regime.}
    \label{fig:selection_by_protocol}
\end{figure}

Figure~\ref{fig:selection_by_protocol} shows that while baseline strategies show high variance and severe performance degradation on specific benchmark splits (indicated by the low-lying scatter points), Mid-Range selection maintains tightly clustered, high-fidelity rankings ($>0.85$ mean $\rho$) regardless of the evaluation regime.

Across all evaluation protocols, the results reveal a clear ordering driven by difficulty filtering. \MR{} achieves the best rank and score prediction ($\rho = 0.910$, $R^2 = 0.670$), closely followed by \textsc{Easiest}-$k$ ($\rho = 0.894$, $R^2 = 0.628$). \textsc{Stratified}-$k$ ($\rho = 0.875$) falls between: better than \Random{} ($\rho = 0.816$) by guaranteeing a balanced sample, but worse than \MR{} because it forces inclusion of hard, noisy tasks. \Greedy{}-$k$ ($\rho = 0.842$) systematically overfits and thus has less reliable performance under temporal and scaffold shift. \textsc{Hardest}-$k$ is catastrophic ($\rho = 0.638$, $R^2 = -3.360$): when agents universally fail selected tasks, discrimination becomes impossible.

The critical insight is that the dominant factor is \emph{which tasks are excluded}, not which are included. Both \MR{} and \textsc{Easiest}-$k$ succeed because they drop the hardest tasks, which contribute noise rather than signal. \MR{} is preferred over \textsc{Easiest}-$k$ for two reasons: it achieves better $R^2$ (by also excluding ceiling-effect tasks where all agents succeed) and features principled theoretical motivation from IRT. \textsc{Stratified}-$k$'s intermediate performance confirms this: forcing inclusion of hard tasks proportional to the original benchmark actively degrades ranking prediction.

Across all evaluation protocols, \MR{} achieves the highest ranking preservation on five out of eight benchmarks and comes in a close second on the rest. \Random{} task selection performs well on average, but this average is computed over 100 seeds, which does not reflect real benchmark practice and would itself defeat the purpose of cost reduction. Its large variance makes it unusable as a deployment strategy; we retain it as a useful baseline.

\begin{table}[H]
\centering
\small
\begin{tabular}{lrcccccr}
\toprule
 & & & & & & \multicolumn{2}{c}{\Random{}} \\
\cmidrule(lr){7-8}
Benchmark & $k$ & \MR{} & \Greedy{} & \Easiest{} & \Stratified{} & $\mu \pm \sigma$ & Worst / Best \\
\midrule
TerminalBench & 39 & 0.980 & \textbf{0.988} & 0.970 & 0.979 & 0.980 $\pm$ .003 & 0.962 / 0.991 \\
GAIA & 77 & \textbf{0.946} & 0.927 & 0.928 & 0.930 & 0.938 $\pm$ .015 & 0.872 / 0.987 \\
SWE-bench Mini & 20 & 0.920 & 0.881 & \textbf{0.922} & 0.902 & 0.903 $\pm$ .026 & 0.695 / 0.977 \\
USACO & 81 & \textbf{0.938} & 0.886 & 0.861 & 0.914 & 0.924 $\pm$ .033 & 0.700 / 0.995 \\
$\tau$-bench Airline & 27 & \textbf{0.944} & 0.834 & 0.920 & 0.874 & 0.860 $\pm$ .038 & 0.708 / 0.957 \\
CoreBench Hard & 14 & \textbf{0.901} & 0.869 & 0.896 & 0.741 & 0.780 $\pm$ .050 & 0.588 / 0.910 \\
Online Mind2Web & 86 & \textbf{0.921} & 0.612 & 0.913 & 0.798 & 0.808 $\pm$ .062 & 0.543 / 0.962 \\
SciCode & 4 & 0.736 & \textbf{0.753} & 0.736 & --- & 0.364 $\pm$ .147 & $-$0.189 / 0.734 \\
\bottomrule
\end{tabular}
\caption{Per-benchmark mean $\rho$ across all evaluation protocols. \Random{} reports mean $\pm$ std over 100 seeds per protocol; Worst/Best denote extreme seed outcomes across all protocols.}
\label{tab:per_benchmark}
\end{table}

\begin{examplebox}{Why \Easiest{} Is Competitive: It Selects Mid-Range Tasks}

The \Easiest{}-$k$ achieves surprisingly competitive ranking performance (Table~\ref{tab:per_benchmark}), but not because easy tasks are inherently informative. On 4 of 8 benchmarks, \Easiest{}-$k$ and \MR{} select identical or near-identical task sets: when a benchmark's difficulty distribution is left-skewed---many hard tasks, few easy ones---the $k$ easiest tasks simply have a higher overlap with the mid-range band. The performance gap between \MR{} and \Easiest{}-$k$ correlates strongly with their task-set overlap ($r = -0.71$, $p = 0.05$): where overlap is high (\textsc{CoreBench}, \textsc{SciCode}, $\tau$-bench), the two methods are nearly indistinguishable; where overlap is low (\textsc{USACO}, 14\% overlap), \MR{} outperforms \Easiest{}-$k$ by $\Delta\rho = 0.078$. \Easiest{}-$k$ is competitive precisely when, and because, it is a noisy approximation of \MR{}.

Appendix~\ref{app:easiest_overlap_distributions} visualizes the
underlying difficulty distributions that drive this overlap.

\begin{figure}[H]
    \centering
    \includegraphics[width=0.9\textwidth]{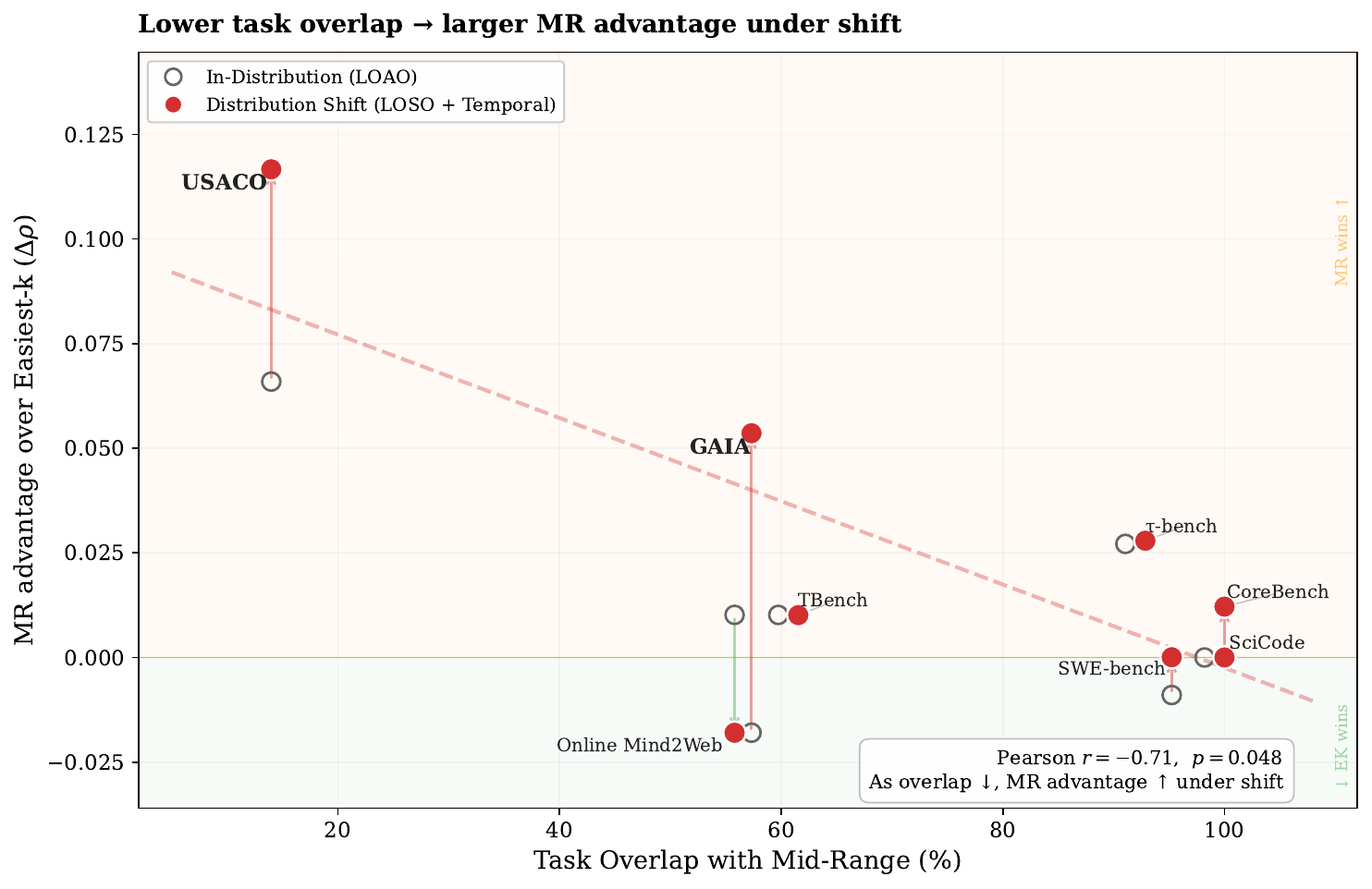}
    \caption{The \MR{}--\Easiest{} performance gap correlates with task-set overlap ($r = -0.71$, $p = 0.048$): \Easiest{}-$k$ is competitive when it selects the same tasks as \MR{}. USACO (14\% overlap) is the benchmark where the two sets genuinely differ, and it is where \MR{} most clearly wins ($\Delta\rho = 0.078$ under distribution shift).}
    \label{fig:easiest_overlap}
\end{figure}

\end{examplebox}

\subsection{Cost Reduction}

Table~\ref{tab:reduction} reports the per-benchmark task and cost reduction from \MR{} filtering. Typical task reduction ranges from 44\% to 70\%, with a median of 58\%. These are significant reductions that translate into per-run savings ranging from \$5--\$22 for $\tau$-bench Airline to \$149--\$253 for Online Mind2Web.

\begin{table}[H]
\centering
\small
\begin{tabular}{lrrr|ccc|rr}
\toprule
 & & & & \multicolumn{3}{c|}{\MR{} $\rho$} & \multicolumn{2}{c}{Cost Savings} \\
\cmidrule(lr){5-7} \cmidrule(lr){8-9}
Benchmark & $N$ & $k_{\mathrm{MR}}$ & Reduction & LOAO & LOSO & Temp. & Median & Range \\
\midrule
TerminalBench & 89 & 39 & 56\% & 0.986 & 0.984 & 0.975 & --- & --- \\
GAIA & 165 & 82 & 50\% & 0.956 & 0.926 & 0.974 & \$70.3 & \$40.1--\$237.7 \\
SWE-bench Mini & 50 & 21 & 58\% & 0.924 & 0.896 & 0.968 & \$94.5 & \$37.9--\$234.0 \\
USACO & 307 & 93 & 70\% & 0.940 & --- & 0.950 & \$39.3 & \$19.6--\$55.8 \\
$\tau$-bench Airline & 50 & 28 & 44\% & 0.967 & 0.956 & 0.945 & \$9.8 & \$5.2--\$22.4 \\
CoreBench Hard & 45 & 15 & 67\% & 0.918 & 0.912 & 0.903 & \$44.1 & \$16.8--\$67.5 \\
Online Mind2Web & 300 & 95 & 68\% & 0.939 & 0.909 & 0.930 & \$188.7 & \$149.2--\$253.2 \\
\midrule
\textbf{Total/Mean} & 1006 & 373 & 63\% & 0.947 & 0.931 & 0.949 & \$446.7 & \$268.8--\$870.6 \\
\bottomrule
\end{tabular}
\caption{Benchmark reduction from mid-range filtering. Task reduction is the fraction eliminated; cost savings (per agent, per benchmark) assume linear scaling with task count, using median agent cost from HAL leaderboards. TerminalBench lacks HAL-style cost data. USACO LOSO is unavailable because all but one agent use a single scaffold. SciCode excluded due to insufficient mid-range tasks ($k=4$).}
\label{tab:reduction}
\end{table}

\section{Discussion}

\subsection{Practical Protocol for Benchmark Operators}

To take advantage of benchmark reduction using the \MR{} task selection strategy, we recommend a four-phase deployment:

\begin{description}[leftmargin=*,itemsep=3pt]
\item[Phase 0---Cold Start.] Use all available historical agent runs to estimate per-task pass rates; even 5--10 runs provide a usable first estimate, and estimates improve continuously as more agents are added. Tasks near the band boundaries will be estimated with more uncertainty under small agent runs, but misclassification at the boundary has limited consequence given the width of the mid-range band.

\item[Phase 1---Setup.] Select all tasks in the 30--70\% pass rate band. If fewer than 10\% of tasks fall in this band, widen incrementally to 25--75\%, then to the theoretically motivated 15--85\% bound derived from the IRT Fisher information criterion $I(\theta) = p(1-p) \geq \frac{1}{2} I_{\max}$.

\item[Phase 2---Ongoing.] Run new agents on selected tasks only. Report rankings, not absolute scores from the reduced suite. If absolute scores must be reported, report Ridge-predicted full-benchmark scores rather than raw scores on the reduced task set, and accompany them with confidence intervals to reflect prediction uncertainty.

\item[Phase 3---Maintenance.] Fix the task set until evidence suggests otherwise. If score prediction is needed, refit Ridge weights every 5--10 agents. Monitor $\rho$ on occasional full-benchmark validation runs; trigger reselection only if $\rho$ falls below 0.75.
\end{description}

This protocol is inappropriate when: only a few agent runs are expected and cold-start cost cannot be recouped; absolute capability claims are needed; or the difficulty distribution is so skewed that fewer than 10\% of tasks fall in the mid-range band (as in \textsc{SciCode}, where fewer than 5 of 65 tasks are mid-range).

\subsection{A Call for Per-Task Transparency}

This study is possible only because HAL and Terminal-Bench publish per-task, per-agent results---the full performance matrix, not just aggregate scores. Most benchmarks do not.

Even when per-task data nominally exists, it is often distributed across nested JSON logs or multiple API endpoints, requiring substantial engineering effort to reconstruct into a usable format. Data that is publicly available but not readily accessible limits reproducibility and community reuse.

We recommend that benchmark operators publish, alongside any leaderboard, a single flat file (CSV or equivalent) containing one row per agent--task pair with columns for agent identifier, task identifier, outcome, scaffold/harness, model, and submission date. This costs nothing beyond what is already computed during evaluation but enables the community to study evaluation efficiency, detect benchmark saturation, identify redundant tasks, and build reduced evaluation suites. The amortization argument depends on this: historical rollouts become shared training data only if they are shared---and shared in a format that anyone can load in one line of code.

\section{Limitations}

\MR{} requires a populated mid-range difficulty band to function properly. It fails on SciCode, where only about four tasks fall into this band, and more generally, benchmarks with highly skewed difficulty distributions are less compatible with this approach.

There is a non-trivial cold-start cost: approximately five to ten agents must be evaluated on the full benchmark before reduction becomes reliable for incremental use, and closer to fifteen agents are needed for stability across all future comparisons.

For HAL benchmarks, the number of agent runs is relatively small (\( n = 13\text{--}38 \)), with limited scaffold diversity and minimal ability to estimate run-to-run variance. We mitigated this using nested cross-validation and replication across eight benchmarks, but larger and more diverse populations would strengthen the conclusions.

For TerminalBench, the temporal validation window spans approximately 3.5 months (105 days), during which the agent population increased by about 40\%. We observe that task selection quality degrades over time as the population evolves. In particular, large capability jumps, reflected as sudden increases in \MR{} pass rates, can invalidate previously selected tasks more quickly than gradual drift. In such cases, a full-benchmark run is required to confirm results, collect additional data, and potentially redo task selection or declare the benchmark saturated.

All benchmarks in this study use binary or near-binary outcomes per task. Benchmarks with grading that obscure the relationship between per-task outcomes and overall score may behave differently and are not directly covered by these results.

Finally, the 30--70\% difficulty band is a practical heuristic motivated by item response theory and cost reduction targets, rather than a theoretically unique or optimal choice.

\section{Conclusion}

Our results show that reliable AI agent leaderboard rankings do not require full-benchmark evaluation. Across eight benchmarks and multiple distribution-shift regimes, ranking remains substantially more stable than absolute score prediction, making reduced evaluation feasible even when calibration degrades.

This motivates a simple practical rule: evaluate new agents on mid-range tasks only. The \MR{} protocol reduces evaluation cost substantially while preserving leaderboard fidelity more reliably than random or greedy alternatives. Its value is operational rather than universal: it is best suited to benchmarks with a populated intermediate-difficulty band, transparent per-task outcomes, and ongoing agent arrivals that justify amortizing the initial full-benchmark cost.

More broadly, the paper advocates for a shift in the way agent evaluation is framed. Agent benchmarks are often treated as instruments for measuring absolute capability---a difficult-to-attain ideal given construct validity failures and non-representative task distributions---but in practice they are primarily used to compare and rank systems. Once evaluation is aligned with that use case, selective measurement becomes a principled and effective alternative to exhaustive benchmarking.

Reduced evaluation should therefore be the default for routine leaderboard maintenance, with full-benchmark runs reserved for initialization, drift checks, major capability jumps, and saturation analysis.

\paragraph{Reproducibility}
All code, data, and pre-computed results are available at \url{https://github.com/fsndzomga/efficient-benchmarking-ai-agents}. The repository includes the per-task performance matrices for all eight benchmarks we used, evaluation scripts for all five protocols (LOAO, LOSO, temporal, random split, within-scaffold LOAO), and CSV outputs that form the basis of all tables and figures in this paper.

\bibliographystyle{plainnat}
\bibliography{references}

\begin{thebibliography}{29}
\providecommand{\natexlab}[1]{#1}
\providecommand{\url}[1]{\texttt{#1}}
\expandafter\ifx\csname urlstyle\endcsname\relax
  \providecommand{\doi}[1]{doi: #1}\else
  \providecommand{\doi}{doi: \begingroup \urlstyle{rm}\Url}\fi

\bibitem[Aider-AI(2024)]{aider2024}
Aider-AI.
\newblock Aider: Ai pair programming in your terminal, 2024.
\newblock URL \url{https://github.com/Aider-AI/aider}.
\newblock GitHub repository.

\bibitem[Baker(2001)]{baker2001_irt}
Frank~B. Baker.
\newblock \emph{The Basics of Item Response Theory}.
\newblock ERIC Clearinghouse on Assessment and Evaluation, College Park, MD, 2nd edition, 2001.
\newblock ISBN 9781886047037.
\newblock \doi{unspecified}.
\newblock URL \url{https://eric.ed.gov/?id=ED458219}.
\newblock Relevance: Introductory IRT reference underpinning “difficulty/discrimination/information” claims often used to justify mid-range item selection.

\bibitem[Besiroglu et~al.(2024)Besiroglu, Bergerson, Michael, Heim, Luo, and Thompson]{besiroglu2024computedividemachinelearning}
Tamay Besiroglu, Sage~Andrus Bergerson, Amelia Michael, Lennart Heim, Xueyun Luo, and Neil Thompson.
\newblock The compute divide in machine learning: A threat to academic contribution and scrutiny?, 2024.
\newblock URL \url{https://arxiv.org/abs/2401.02452}.

\bibitem[Chen et~al.(2025)Chen, Chen, Ning, Zhang, Wang, Yu, Li, Liao, Wei, Lu, Dey, Xue, Baker, Burns, Adu-Ampratwum, Huang, Ning, Gao, Su, and Sun]{chen2025scienceagentbenchrigorousassessmentlanguage}
Ziru Chen, Shijie Chen, Yuting Ning, Qianheng Zhang, Boshi Wang, Botao Yu, Yifei Li, Zeyi Liao, Chen Wei, Zitong Lu, Vishal Dey, Mingyi Xue, Frazier~N. Baker, Benjamin Burns, Daniel Adu-Ampratwum, Xuhui Huang, Xia Ning, Song Gao, Yu~Su, and Huan Sun.
\newblock Scienceagentbench: Toward rigorous assessment of language agents for data-driven scientific discovery, 2025.
\newblock URL \url{https://arxiv.org/abs/2410.05080}.

\bibitem[Gignac and Ilić(2025)]{GIGNAC2025101922}
Gilles~E. Gignac and David Ilić.
\newblock Psychometrically derived 60-question benchmarks: Substantial efficiencies and the possibility of human-ai comparisons.
\newblock \emph{Intelligence}, 110:\penalty0 101922, 2025.
\newblock ISSN 0160-2896.
\newblock \doi{https://doi.org/10.1016/j.intell.2025.101922}.
\newblock URL \url{https://www.sciencedirect.com/science/article/pii/S016028962500025X}.

\bibitem[Gonzalez et~al.(2025)Gonzalez, Hernandez, Perez, Orozco, Soto, and Malagon]{gonzalez2025repetitionsmatterstrengtheningreliability}
Miguel Angel~Alvarado Gonzalez, Michelle~Bruno Hernandez, Miguel Angel~Peñaloza Perez, Bruno~Lopez Orozco, Jesus Tadeo~Cruz Soto, and Sandra Malagon.
\newblock Do repetitions matter? strengthening reliability in llm evaluations, 2025.
\newblock URL \url{https://arxiv.org/abs/2509.24086}.

\bibitem[Hendrycks et~al.(2021)Hendrycks, Burns, Basart, Zou, Mazeika, Song, and Steinhardt]{hendrycks2021measuring}
Dan Hendrycks, Collin Burns, Steven Basart, Andy Zou, Mantas Mazeika, Dawn Song, and Jacob Steinhardt.
\newblock Measuring massive multitask language understanding.
\newblock In \emph{International Conference on Learning Representations}, 2021.
\newblock URL \url{https://openreview.net/forum?id=d7KBjmI3GmQ}.

\bibitem[Hoerl and Kennard(1970)]{Hoerl01021970}
Arthur~E. Hoerl and Robert~W. Kennard.
\newblock Ridge regression: Biased estimation for nonorthogonal problems.
\newblock \emph{Technometrics}, 12\penalty0 (1):\penalty0 55--67, 1970.
\newblock \doi{10.1080/00401706.1970.10488634}.
\newblock URL \url{https://doi.org/10.1080/00401706.1970.10488634}.

\bibitem[Jimenez et~al.(2024)Jimenez, Yang, Wettig, Yao, Pei, Press, and Narasimhan]{jimenez2024swebenchlanguagemodelsresolve}
Carlos~E. Jimenez, John Yang, Alexander Wettig, Shunyu Yao, Kexin Pei, Ofir Press, and Karthik Narasimhan.
\newblock Swe-bench: Can language models resolve real-world github issues?, 2024.
\newblock URL \url{https://arxiv.org/abs/2310.06770}.

\bibitem[Kapoor et~al.(2026)Kapoor, Stroebl, Kirgis, Nadgir, Siegel, Wei, Xue, Chen, Chen, Utpala, Ndzomga, Oruganty, Luskin, Liu, Yu, Arora, Hahm, Trivedi, Sun, Lee, Jin, Mai, Zhou, Zhu, Bommasani, Kang, Song, Henderson, Su, Liang, and Narayanan]{hal2026}
Sayash Kapoor, Benedikt Stroebl, Peter Kirgis, Nitya Nadgir, Zachary~S Siegel, Boyi Wei, Tianci Xue, Ziru Chen, Felix Chen, Saiteja Utpala, Franck Ndzomga, Dheeraj Oruganty, Sophie Luskin, Kangheng Liu, Botao Yu, Amit Arora, Dongyoon Hahm, Harsh Trivedi, Huan Sun, Juyong Lee, Tengjun Jin, Yifan Mai, Yifei Zhou, Yuxuan Zhu, Rishi Bommasani, Daniel Kang, Dawn Song, Peter Henderson, Yu~Su, Percy Liang, and Arvind Narayanan.
\newblock Holistic agent leaderboard: The missing infrastructure for ai agent evaluation.
\newblock In \emph{The Fourteenth International Conference on Learning Representations}, 2026.
\newblock \doi{10.48550/arXiv.2510.11977}.
\newblock URL \url{https://openreview.net/forum?id=vUaY1t64ZZ}.
\newblock Relevance: Introduces HAL as a standardized, cost-aware, third-party evaluation platform and harness, enabling cross-scaffold agent evaluation analyses central to minimal-suite selection.

\bibitem[Liang et~al.(2023)Liang, Bommasani, Lee, Tsipras, Soylu, Yasunaga, Zhang, Narayanan, Wu, Kumar, Newman, Yuan, Yan, Zhang, Cosgrove, Manning, Ré, Acosta-Navas, Hudson, Zelikman, Durmus, Ladhak, Rong, Ren, Yao, Wang, Santhanam, Orr, Zheng, Yuksekgonul, Suzgun, Kim, Guha, Chatterji, Khattab, Henderson, Huang, Chi, Xie, Santurkar, Ganguli, Hashimoto, Icard, Zhang, Chaudhary, Wang, Li, Mai, Zhang, and Koreeda]{liang2023holisticevaluationlanguagemodels}
Percy Liang, Rishi Bommasani, Tony Lee, Dimitris Tsipras, Dilara Soylu, Michihiro Yasunaga, Yian Zhang, Deepak Narayanan, Yuhuai Wu, Ananya Kumar, Benjamin Newman, Binhang Yuan, Bobby Yan, Ce~Zhang, Christian Cosgrove, Christopher~D. Manning, Christopher Ré, Diana Acosta-Navas, Drew~A. Hudson, Eric Zelikman, Esin Durmus, Faisal Ladhak, Frieda Rong, Hongyu Ren, Huaxiu Yao, Jue Wang, Keshav Santhanam, Laurel Orr, Lucia Zheng, Mert Yuksekgonul, Mirac Suzgun, Nathan Kim, Neel Guha, Niladri Chatterji, Omar Khattab, Peter Henderson, Qian Huang, Ryan Chi, Sang~Michael Xie, Shibani Santurkar, Surya Ganguli, Tatsunori Hashimoto, Thomas Icard, Tianyi Zhang, Vishrav Chaudhary, William Wang, Xuechen Li, Yifan Mai, Yuhui Zhang, and Yuta Koreeda.
\newblock Holistic evaluation of language models, 2023.
\newblock URL \url{https://arxiv.org/abs/2211.09110}.

\bibitem[Merrill et~al.(2026)]{terminalbench2026}
Mike~A. Merrill et~al.
\newblock Terminal-bench: Benchmarking agents on hard, realistic tasks in command line interfaces.
\newblock arXiv preprint, 2026.
\newblock URL \url{https://arxiv.org/abs/2601.11868}.
\newblock Relevance: Independent agent benchmark ecosystem (terminal/CLI) used for out-of-distribution validation of minimal evaluation suites and cross-scaffold transfer claims.

\bibitem[Mialon et~al.(2024)Mialon, Fourrier, Wolf, LeCun, and Scialom]{gaia2024}
Gr{\'e}goire Mialon, Cl{\'e}mentine Fourrier, Thomas Wolf, Yann LeCun, and Thomas Scialom.
\newblock Gaia: a benchmark for general ai assistants.
\newblock In \emph{The Twelfth International Conference on Learning Representations}, 2024.
\newblock \doi{10.48550/arXiv.2311.12983}.
\newblock URL \url{https://openreview.net/forum?id=fibxvahvs3}.
\newblock Relevance: Tool-using assistant benchmark; exemplifies expensive, heterogeneous agent evaluation motivating subset-based cost reductions.

\bibitem[OpenAI(2025)]{openai_codex_cli_2025}
OpenAI.
\newblock Codex: Lightweight coding agent for the terminal, 2025.
\newblock URL \url{https://github.com/openai/codex}.
\newblock GitHub repository.

\bibitem[Perlitz et~al.(2024)Perlitz, Bandel, Gera, Arviv, Ein-Dor, Shnarch, Slonim, Shmueli-Scheuer, and Choshen]{perlitz2024efficientbenchmarkinglanguagemodels}
Yotam Perlitz, Elron Bandel, Ariel Gera, Ofir Arviv, Liat Ein-Dor, Eyal Shnarch, Noam Slonim, Michal Shmueli-Scheuer, and Leshem Choshen.
\newblock Efficient benchmarking of language models, 2024.
\newblock URL \url{https://arxiv.org/abs/2308.11696}.

\bibitem[Polo et~al.(2024)Polo, Weber, Choshen, Sun, Xu, and Yurochkin]{tinybench2024}
Felipe~Maia Polo, Lucas Weber, Leshem Choshen, Yuekai Sun, Gongjun Xu, and Mikhail Yurochkin.
\newblock tinybenchmarks: evaluating llms with fewer examples.
\newblock arXiv preprint, 2024.
\newblock URL \url{https://arxiv.org/abs/2402.14992}.
\newblock Relevance: Demonstrates that carefully selected tiny subsets can approximate full-benchmark ranking/score estimates for LLMs; directly analogous to “minimal suites” for agents.

\bibitem[Rodriguez et~al.(2021)Rodriguez, Barrow, Hoyle, Lalor, Jia, and Boyd-Graber]{rodriguez-etal-2021-evaluation}
Pedro Rodriguez, Joe Barrow, Alexander Hoyle, John~P. Lalor, Robin Jia, and Jordan Boyd-Graber.
\newblock Evaluation examples are not equally informative: How should that change {NLP} leaderboards?
\newblock In Chengqing Zong, Fei Xia, Wenjie Li, and Roberto Navigli, editors, \emph{Proceedings of the 59th Annual Meeting of the Association for Computational Linguistics and the 11th International Joint Conference on Natural Language Processing (Volume 1: Long Papers)}, pages 4486--4503, Online, August 2021. Association for Computational Linguistics.
\newblock \doi{10.18653/v1/2021.acl-long.346}.
\newblock URL \url{https://aclanthology.org/2021.acl-long.346/}.

\bibitem[Shi et~al.(2024)Shi, Tang, Narasimhan, and Yao]{shi2024languagemodelssolveolympiad}
Quan Shi, Michael Tang, Karthik Narasimhan, and Shunyu Yao.
\newblock Can language models solve olympiad programming?, 2024.
\newblock URL \url{https://arxiv.org/abs/2404.10952}.

\bibitem[Siegel et~al.(2024)Siegel, Kapoor, Nagdir, Stroebl, and Narayanan]{siegel2024corebenchfosteringcredibilitypublished}
Zachary~S. Siegel, Sayash Kapoor, Nitya Nagdir, Benedikt Stroebl, and Arvind Narayanan.
\newblock Core-bench: Fostering the credibility of published research through a computational reproducibility agent benchmark, 2024.
\newblock URL \url{https://arxiv.org/abs/2409.11363}.

\bibitem[Srivastava et~al.(2023)]{srivastava2023beyond}
Aarohi Srivastava et~al.
\newblock Beyond the imitation game: Quantifying and extrapolating the capabilities of language models.
\newblock \emph{Transactions on Machine Learning Research}, 2023.
\newblock ISSN 2835-8856.
\newblock URL \url{https://openreview.net/forum?id=uyTL5Bvosj}.
\newblock Featured Certification.

\bibitem[Subramani et~al.(2025)Subramani, Gomez, and Diab]{subramani2025simbasimplifyingbenchmarkanalysis}
Nishant Subramani, Alfredo Gomez, and Mona Diab.
\newblock Simba: Simplifying benchmark analysis using performance matrices alone, 2025.
\newblock URL \url{https://arxiv.org/abs/2510.17998}.

\bibitem[Tian et~al.(2024)Tian, Gao, Zhang, Chen, Fan, Guo, Haas, Ji, Krongchon, Li, Liu, Luo, Ma, Tong, Trinh, Tian, Wang, Wu, Xiong, Yin, Zhu, Lieret, Lu, Liu, Du, Tao, Press, Callan, Huerta, and Peng]{tian2024scicoderesearchcodingbenchmark}
Minyang Tian, Luyu Gao, Shizhuo~Dylan Zhang, Xinan Chen, Cunwei Fan, Xuefei Guo, Roland Haas, Pan Ji, Kittithat Krongchon, Yao Li, Shengyan Liu, Di~Luo, Yutao Ma, Hao Tong, Kha Trinh, Chenyu Tian, Zihan Wang, Bohao Wu, Yanyu Xiong, Shengzhu Yin, Minhui Zhu, Kilian Lieret, Yanxin Lu, Genglin Liu, Yufeng Du, Tianhua Tao, Ofir Press, Jamie Callan, Eliu Huerta, and Hao Peng.
\newblock Scicode: A research coding benchmark curated by scientists, 2024.
\newblock URL \url{https://arxiv.org/abs/2407.13168}.

\bibitem[Vivek et~al.(2024)Vivek, Ethayarajh, Yang, and Kiela]{vivek2024anchorpointsbenchmarkingmodels}
Rajan Vivek, Kawin Ethayarajh, Diyi Yang, and Douwe Kiela.
\newblock Anchor points: Benchmarking models with much fewer examples, 2024.
\newblock URL \url{https://arxiv.org/abs/2309.08638}.

\bibitem[Wang et~al.(2025)Wang, Li, Song, Xu, Tang, Zhuge, Pan, Song, Li, Singh, Tran, Li, Ma, Zheng, Qian, Shao, Muennighoff, Zhang, Hui, Lin, Brennan, Peng, Ji, and Neubig]{wang2025openhandsopenplatformai}
Xingyao Wang, Boxuan Li, Yufan Song, Frank~F. Xu, Xiangru Tang, Mingchen Zhuge, Jiayi Pan, Yueqi Song, Bowen Li, Jaskirat Singh, Hoang~H. Tran, Fuqiang Li, Ren Ma, Mingzhang Zheng, Bill Qian, Yanjun Shao, Niklas Muennighoff, Yizhe Zhang, Binyuan Hui, Junyang Lin, Robert Brennan, Hao Peng, Heng Ji, and Graham Neubig.
\newblock Openhands: An open platform for ai software developers as generalist agents, 2025.
\newblock URL \url{https://arxiv.org/abs/2407.16741}.

\bibitem[Xue et~al.(2025)Xue, Qi, Shi, Song, Gou, Song, Sun, and Su]{xue2025illusionprogressassessingcurrent}
Tianci Xue, Weijian Qi, Tianneng Shi, Chan~Hee Song, Boyu Gou, Dawn Song, Huan Sun, and Yu~Su.
\newblock An illusion of progress? assessing the current state of web agents, 2025.
\newblock URL \url{https://arxiv.org/abs/2504.01382}.

\bibitem[Yao et~al.(2025)Yao, Shinn, Razavi, and Narasimhan]{yao2025taubench}
Shunyu Yao, Noah Shinn, Pedram Razavi, and Karthik~R Narasimhan.
\newblock \{\${\textbackslash}tau\$\}-bench: A benchmark for {\textbackslash}underline\{T\}ool-{\textbackslash}underline\{A\}gent-{\textbackslash}underline\{U\}ser interaction in real-world domains.
\newblock In \emph{The Thirteenth International Conference on Learning Representations}, 2025.
\newblock URL \url{https://openreview.net/forum?id=roNSXZpUDN}.

\bibitem[Yoran et~al.(2024)Yoran, Amouyal, Malaviya, Bogin, Press, and Berant]{yoran2024assistantbenchwebagentssolve}
Ori Yoran, Samuel~Joseph Amouyal, Chaitanya Malaviya, Ben Bogin, Ofir Press, and Jonathan Berant.
\newblock Assistantbench: Can web agents solve realistic and time-consuming tasks?, 2024.
\newblock URL \url{https://arxiv.org/abs/2407.15711}.

\bibitem[Zhou et~al.(2026)Zhou, Huang, Zhao, Han, Wang, Chen, Yang, Bao, Dong, Xu, Zhu, Cao, and Zhao]{zhou2026lostbenchmarksrethinkinglarge}
Hongli Zhou, Hui Huang, Ziqing Zhao, Lvyuan Han, Huicheng Wang, Kehai Chen, Muyun Yang, Wei Bao, Jian Dong, Bing Xu, Conghui Zhu, Hailong Cao, and Tiejun Zhao.
\newblock Lost in benchmarks? rethinking large language model benchmarking with item response theory, 2026.
\newblock URL \url{https://arxiv.org/abs/2505.15055}.

\bibitem[Zhu et~al.(2025)Zhu, Jin, Pruksachatkun, Zhang, Liu, Cui, Kapoor, Longpre, Meng, Weiss, Barez, Gupta, Dhamala, Merizian, Giulianelli, Coppock, Ududec, Sekhon, Steinhardt, Kellermann, Schwettmann, Zaharia, Stoica, Liang, and Kang]{zhu2025establishingbestpracticesbuilding}
Yuxuan Zhu, Tengjun Jin, Yada Pruksachatkun, Andy Zhang, Shu Liu, Sasha Cui, Sayash Kapoor, Shayne Longpre, Kevin Meng, Rebecca Weiss, Fazl Barez, Rahul Gupta, Jwala Dhamala, Jacob Merizian, Mario Giulianelli, Harry Coppock, Cozmin Ududec, Jasjeet Sekhon, Jacob Steinhardt, Antony Kellermann, Sarah Schwettmann, Matei Zaharia, Ion Stoica, Percy Liang, and Daniel Kang.
\newblock Establishing best practices for building rigorous agentic benchmarks, 2025.
\newblock URL \url{https://arxiv.org/abs/2507.02825}.

\end{thebibliography}

\clearpage
\appendix
\section{Appendix}

\paragraph{Use of Large Language Models.} LLMs were used for text editing, proofreading, chart polishing, and coding assistance during the preparation of this work. The authors assume full responsibility for all text and results presented in this paper.

\subsection{A Theory of Why Rankings Are Easier Than Scores}
\label{sec:theory}

This section provides a possible theoretical basis for the empirical results in Section~\ref{sec:results}. We characterize why rank prediction from a task subset is more robust than score prediction, even under scaffold and temporal shift.

\subsubsection{Two Sources of Distortion}

The reduced evaluation pipeline introduces two transformations between the quantity we observe and the quantity we want to recover. First, \emph{subset restriction}: we observe performance on a subset $S$ of tasks rather than the full benchmark. Second, \emph{scaffold shift}: the agents we evaluate may use scaffolds not represented in the historical data used to select $S$. Reliable ranking from a reduced suite requires that the composition of these two transformations be approximately monotone. A simple way to formalize this is through affine distortions.

Let $y(m)$ denote agent $m$'s full-benchmark score and $\bar{x}_S(m)$ its mean score on the selected subset $S$. Even without scaffold shift, subset restriction introduces a distortion:
\begin{equation}
  y(m) \approx \alpha_{\text{sub}} \cdot \bar{x}_S(m) + \beta_{\text{sub}} + \varepsilon_{\text{sub}}(m)
\end{equation}
where $\alpha_{\text{sub}} > 0$ captures the scale mismatch between subset and full-benchmark difficulty, $\beta_{\text{sub}}$ is a level shift, and $\varepsilon_{\text{sub}}(m)$ is a residual. When $\alpha_{\text{sub}} > 0$ and the residual is small, the mapping is approximately monotone: the subset recovers rankings even if it does not recover calibrated scores. The mid-range difficulty filter is designed to keep this residual small. Tasks near 50\% pass rate carry maximal Fisher information about latent agent ability under an IRT model, so the subset captures the region of the difficulty spectrum where agents are most reliably discriminated. By contrast, very hard tasks (where most agents fail) and very easy tasks (where most agents succeed) contribute little to the ordering signal, and their exclusion has limited effect on the monotonicity of the subset-to-full mapping.

\subsubsection{Scaffold Shift and Approximate Monotonicity}

Now consider the second transformation. When a new scaffold $S_2$ appears that was not present during task selection, the relationship between subset scores under the original scaffolds and full-benchmark scores under the new scaffold involves a further distortion:
\begin{equation}
  y_{S_2}(m) \approx \alpha \cdot \bar{x}_{S}(m) + \beta + \varepsilon(m)
\end{equation}
where $y_{S_2}(m)$ is agent $m$'s full-benchmark score under scaffold $S_2$, and $\bar{x}_{S}(m)$ is computed from historical scaffolds. This composed transformation---subset restriction followed by scaffold shift---preserves rankings as long as it remains approximately monotone, meaning $\alpha > 0$ and $\varepsilon(m)$ is small enough not to invert pairwise orderings. Score prediction is more demanding: it requires recovering the correct scale ($\alpha \approx 1$) and intercept ($\beta \approx 0$), which is sensitive to any shift in these parameters.

This framing would explain the $\rho$--$R^2$ divergence observed in Section~\ref{sec:results}. The LOSO protocol measures the composed distortion directly: Spearman $\rho$ remains above 0.90 even when $R^2$ drops below 0.65, confirming that the composition of subset restriction and scaffold shift is approximately rank-preserving in practice, even when it is not score-preserving.

Note that approximate monotonicity under scaffold shift does not require scaffold changes to be model-independent in a strong sense. Some scaffolds are co-optimized with specific model families (e.g., through system prompts, tool schemas, or post-training), which introduces genuine scaffold--model interactions. Two factors may explain why rankings nonetheless remain stable in aggregate. First, scaffold engineering has practical limits: it is difficult to design a harness that boosts a weaker model's performance so dramatically that it overtakes a substantially stronger one on a sufficiently challenging benchmark. Co-optimized scaffolds can narrow the gap between models, but the improvement they provide is typically insufficient to reverse the ordering imposed by underlying capability differences, if these capability differences are high enough. Second, co-optimized scaffolds may represent a minority of the scaffold population: if most scaffolds are approximately model-agnostic, the few with strong model interactions are diluted when averaging over the full agent set. These explanations are not mutually exclusive, and our data do not distinguish between them. What the empirical results establish is that, regardless of mechanism, the composed transformation remains approximately monotone at the population level observed across all 8 benchmarks we studied.

\subsubsection{Temporal Shift as Composed Transformations}

Temporal shift---the distribution change that occurs as new agents are submitted over time---layers a third source of distortion on top of subset restriction and scaffold shift. Over the 3.5-month temporal window in Terminal-Bench, the agent population doubled with 17 novel scaffold types and newer foundation models. Each new scaffold introduces its own approximately monotone distortion relative to the historical data used for task selection. A composition of approximately monotone functions is itself approximately monotone, so the cumulative temporal shift inherits the rank-preserving property of its components, provided no individual shift is severe enough to invert the ordering.

The capability component of temporal shift---new, more powerful foundation models---could in principle break this structure. A sufficiently large capability jump would constitute a nonlinear shift that violates the monotone approximation: if an entire class of previously intractable tasks becomes routine, the mid-range band estimated from historical data would no longer target the discriminative region. However, the temporal expanding-window protocol achieves $\rho = 0.921$, only 0.01 below the LOAO baseline, suggesting that capability improvements during the observation period were incremental enough to remain within the monotone regime. This is not guaranteed to hold indefinitely, which is why the deployment protocol in Section~5.1 prescribes periodic full-benchmark runs when evidence of drift or discontinuity appears.

\subsection{Post-Hoc Verification of Mid-Range Band Selection}
\label{app:band_sensitivity}

As stated previously, for the Mid-Range task selection strategy, we filter tasks to those with historical pass rates in the interval $[0.30, 0.70]$ (the 30--70 band). It is crucial to emphasize that this specific threshold was selected \textit{a priori}. The choice was motivated by principles from Item Response Theory (IRT)---which posits that items with pass rates near $p=0.5$ provide maximal discriminative power between closely matched capabilities---balanced against a practical desire for more aggressive computational cost reduction than theoretically broader optimal intervals (e.g., 15--85). Choosing the filtering band by optimizing for correlation directly on the benchmark data would introduce circularity and overstate the generalizability of the subset selection.

To verify that our \textit{a priori} selection was reasonable, we performed a post-hoc sensitivity analysis across seven benchmarks (excluding Scicode due to data size constraints). We evaluated how tightening the percentile band affects both the rank preservation (measured by Spearman $\rho$ against the full benchmark) and the total task reduction (computational savings).

\begin{figure}[htbp]
    \centering
    % Adjust the path to wherever you store your figures
    \includegraphics[width=0.7\linewidth]{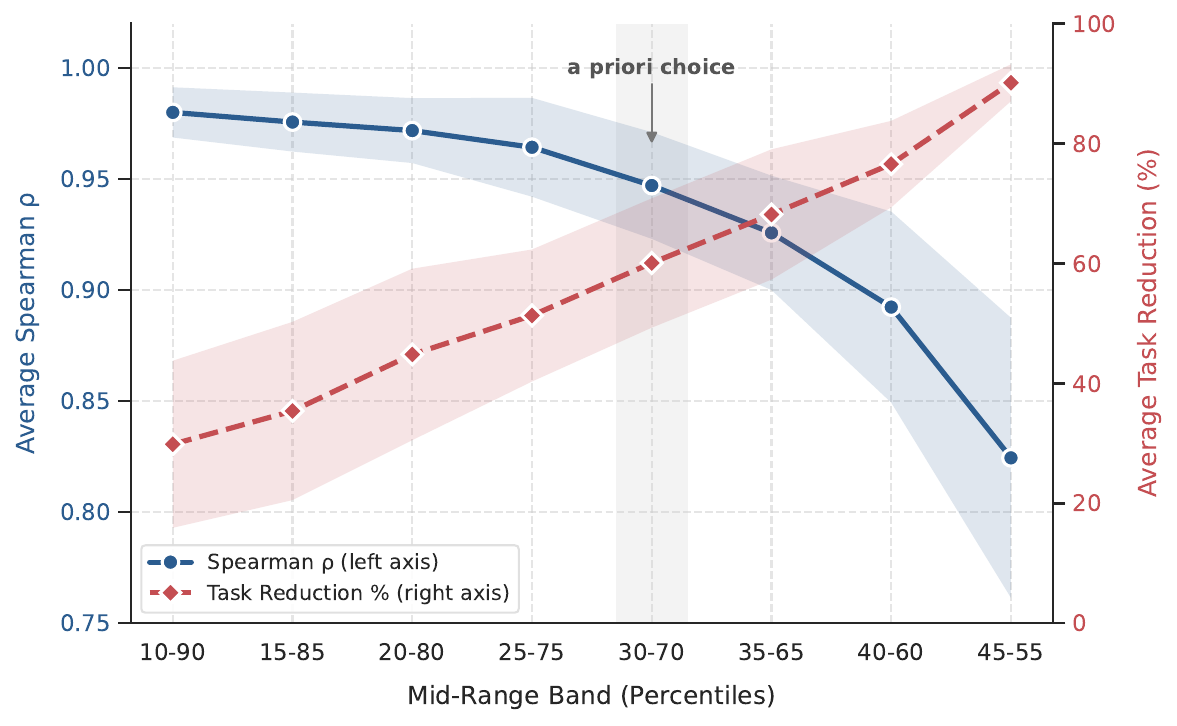}
    \caption{\textbf{Post-hoc sensitivity analysis of mid-range bands.} The average Spearman rank correlation $\rho$ (blue, left axis) and average task reduction percentage (red, right axis) across seven benchmarks for increasingly narrow mid-range bands. Shaded regions denote $\pm 1$ standard deviation across the benchmarks. The 30--70 band, chosen \textit{a priori}, strikes a highly favorable balance.}
    \label{fig:band_sensitivity}
\end{figure}

As shown in \Cref{fig:band_sensitivity}, there is a clear empirical trade-off between rank preservation and task reduction. Wider bands (e.g., 10--90) achieve near-perfect correlation ($\rho \approx 0.98$) but only reduce tasks by roughly $30\%$. Extremely tight bands (e.g., 45--55) reduce tasks by $90\%$ but see correlations drop more precipitously and with higher variance across benchmarks.

The pre-selected 30--70 band occupies an effective position on this Pareto frontier. Averaged across the evaluated benchmarks, this band maintains a highly reliable Spearman correlation of $\rho = 0.95 \pm 0.02$, while yielding a substantial average task reduction of $60\% \pm 11\%$. This post-hoc analysis confirms that our theoretically motivated, \textit{a priori} choice provides robust, high-fidelity rankings without requiring data-snooping optimizations.

\subsection{Per-Benchmark Performance by Task Selection Strategy}
\label{app:per_benchmark_heatmap}
 
\begin{figure}[htbp]
    \centering
    \includegraphics[width=\textwidth]{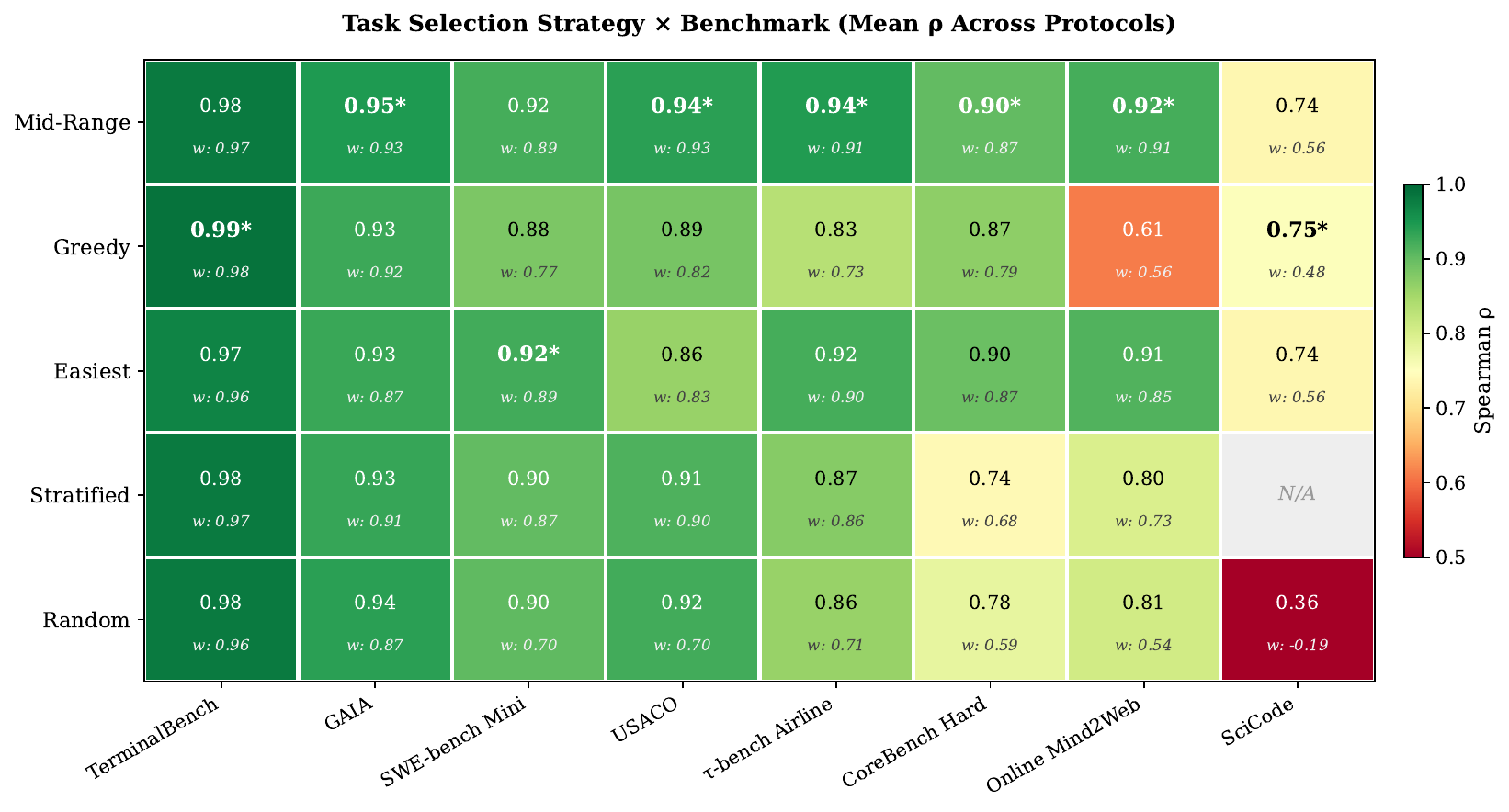}
    \caption{\textbf{Task Selection Strategy $\times$ Benchmark (Mean $\rho$ Across Protocols).} Each cell shows the mean Spearman~$\rho$ across evaluation protocols; worst-case $\rho$ is shown below in gray. Asterisks mark the best-performing strategy per benchmark.}
    \label{fig:heatmap_strategy_benchmark}
\end{figure}

Figure~\ref{fig:heatmap_strategy_benchmark} reports mean Spearman~$\rho$ (averaged across evaluation protocols) for each combination of task selection strategy and benchmark, along with worst-case $\rho$ in gray.
 
Several patterns emerge from the per-benchmark breakdown.
 
\MR{} achieves the best or tied-best mean $\rho$ on five of eight
benchmarks (GAIA, USACO, $\tau$-bench Airline, CoreBench Hard, Online Mind2Web) and is within 0.01 of the best on the remaining three. More importantly, its worst-case performance never drops below 0.56 (SciCode, where the mid-range band contains only a handful of tasks), and stays above 0.87 on all benchmarks where the protocol is well-defined.
 
\Greedy{} achieves the highest single-benchmark $\rho$ (0.99 on
Terminal-Bench) but pays for this with poor worst cases: it drops to 0.48 on SciCode and 0.56 on Online Mind2Web, confirming that
data-driven selection overfits when the agent population shifts.
 
\Easiest{} is competitive on benchmarks with left-skewed difficulty distributions (CoreBench Hard, $\tau$-bench Airline), where the easiest tasks overlap substantially with the mid-range band.  Where the two task sets genuinely diverge---most visibly on USACO---\MR{} pulls ahead.
 
\Random{} shows the characteristic pattern discussed in the main text: respectable means but worst cases as low as $-0.19$ (SciCode) and 0.54 (Online Mind2Web), making it unusable as a deployment rule despite acceptable average performance. Also, random selection is averaged over 100 seeds, which itself defeats the purpose of cost reduction in practice.
 
SciCode is an outlier across all strategies. Its difficulty distribution is heavily skewed, leaving very few tasks in the mid-range band. All strategies degrade on this benchmark, which confirms the limitation noted in Section~5: \MR{} requires a sufficiently populated intermediate-difficulty region to function reliably.

\subsection{Task Difficulty Distributions: Easiest-$k$ vs.\ Mid-Range Overlap}
\label{app:easiest_overlap_distributions}
 
\begin{figure}[htbp]
    \centering
    \includegraphics[width=\textwidth]{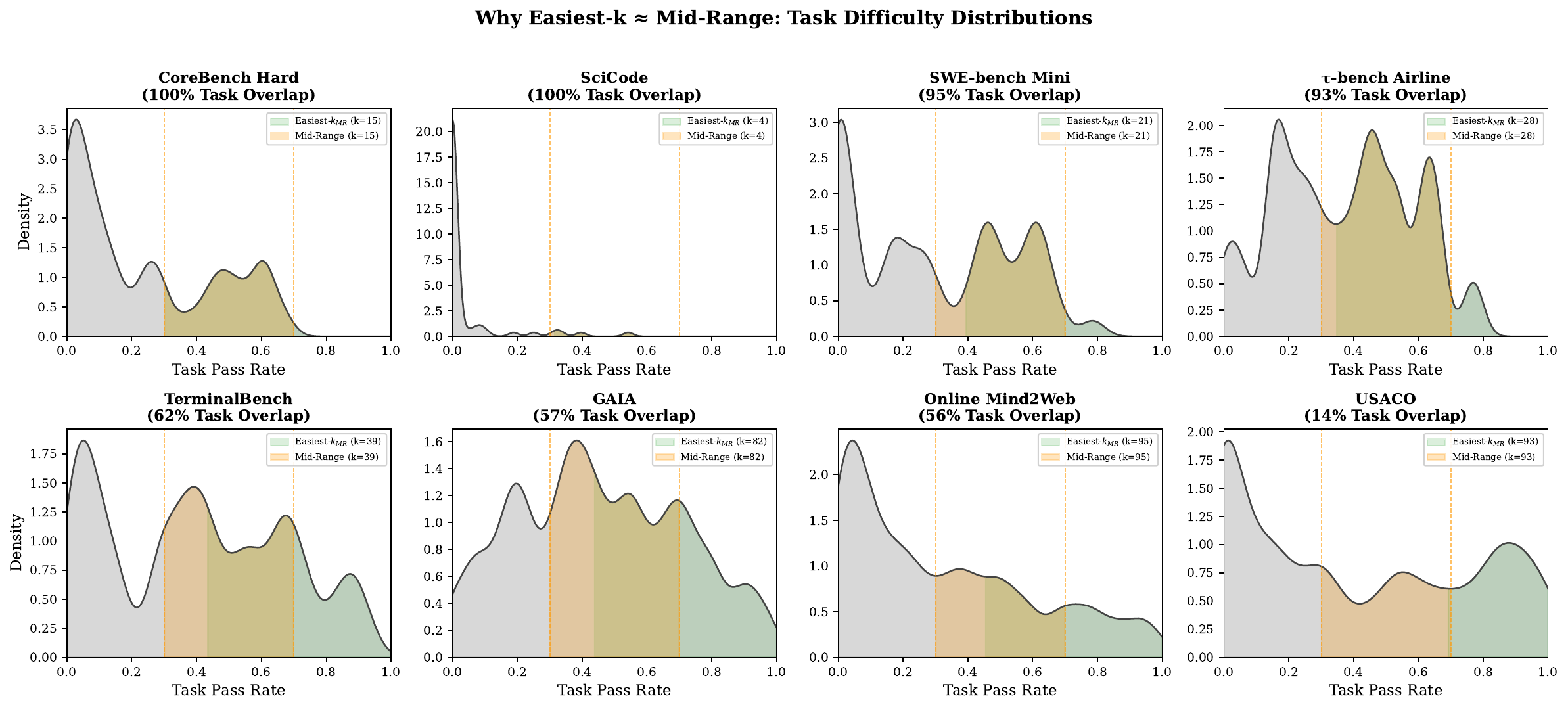}
    \caption{\textbf{Why \Easiest{}-$k$ $\approx$ \MR{}: Task Difficulty Distributions.} Each panel shows the task pass-rate density for one benchmark, with \Easiest{}-$k$ (green) and \MR{} (orange) task sets overlaid. Dashed lines mark the 30--70\% mid-range band. Panels are ordered by decreasing task overlap between the two strategies.}
    \label{fig:easiest_mr_distributions}
\end{figure}

Figure~\ref{fig:easiest_mr_distributions} shows the task difficulty distribution for each benchmark, with the \Easiest{}-$k$ and \MR{} task sets highlighted. The vertical dashed lines mark the 30--70\% mid-range band.
 
The figure makes visible the mechanism behind \Easiest{}-$k$'s
surprisingly competitive performance. When a benchmark's difficulty distribution is left-skewed---concentrated toward hard tasks with a thinner right tail---the $k$ easiest tasks fall inside or near the mid-range band, producing high overlap and near-identical ranking fidelity. CoreBench Hard and SciCode reach 100\% overlap: every task selected by \Easiest{}-$k$ is also a mid-range task. SWE-bench Mini (95\%) and $\tau$-bench Airline (93\%) are close behind.
 
As overlap decreases, the two strategies diverge. TerminalBench (62\%), GAIA (57\%), and Online Mind2Web (56\%) show moderate separation: \Easiest{}-$k$ begins pulling in ceiling-effect tasks where most agents succeed, which contribute less discriminative signal. USACO is the clearest case of genuine divergence at 14\% overlap. Its difficulty distribution is right-skewed, so the easiest tasks cluster near pass rates of 0.7--1.0, well above the mid-range band. This is exactly the benchmark where \MR{} most clearly outperforms \Easiest{}-$k$ ($\Delta\rho = 0.078$ under distribution shift), confirming that the performance gap tracks task-set divergence rather than any intrinsic advantage of easy tasks.

\end{document}